\documentclass[a4paper,conference]{IEEEtran}

\usepackage{ltexpprt}
\usepackage[utf8x]{inputenc}
\usepackage{amsmath, amssymb, amsfonts}
\usepackage{algorithm, algpseudocode}
\usepackage[dvipsnames]{xcolor}
\usepackage{times, enumitem, booktabs, newtxmath, graphicx, textcomp, bbm, cite, url, scalerel} 
\usepackage[colorlinks, linkcolor=blue, urlcolor=blue, citecolor=blue, anchorcolor=blue]{hyperref}

\DeclareMathOperator*{\argmin}{arg\,min}
\newcommand{\etal}{\textit{et~al.}}
\newcommand{\bX}{\boldsymbol{X}}
\newcommand{\bA}{\boldsymbol{A}}
\newcommand{\bB}{\boldsymbol{B}}
\newcommand{\bx}{\boldsymbol{x}}
\newcommand{\by}{\boldsymbol{y}}
\newcommand{\bv}{\boldsymbol{v}}

\newcommand{\bbeta}{\boldsymbol{\beta}}

\newcommand{\bSigma}{\boldsymbol{\Sigma}}

\begin{document}

\title{Interpretable Gait Recognition by Granger Causality}
\author{
\vspace{-10pt}
\IEEEauthorblockN{
Michal Balazia\IEEEauthorrefmark{1}\IEEEauthorrefmark{3},
Kate\v{r}ina Hlav\'a\v{c}kov\'{a}-Schindler\IEEEauthorrefmark{2}\IEEEauthorrefmark{4},
Petr Sojka\IEEEauthorrefmark{3},
Claudia Plant\IEEEauthorrefmark{2}
}
\IEEEauthorblockA{\small\IEEEauthorrefmark{1}INRIA Sophia Antipolis - M\'{e}diterran\'{e}e, 2004 Route des Lucioles, 06902 Sophia Antipolis, France}
\vspace{-17pt}
\IEEEauthorblockA{\small\IEEEauthorrefmark{2}Faculty of Computer Science, University of Vienna, W\"{a}hringer Strasse 29, 1090 Vienna, Austria}
\vspace{-17pt}
\IEEEauthorblockA{\small\IEEEauthorrefmark{3}Faculty of Informatics, Masaryk University, Botanick\'{a} 68a, 60200 Brno, Czech Republic}
\vspace{-17pt}
\IEEEauthorblockA{\small\IEEEauthorrefmark{4}Institute of Computer Science, Czech Academy of Sciences, Pod Vod\'{a}renskou v\v{e}\v{z}\'{i} 271/2, 18207 Prague, Czech Republic}
\vspace{-17pt}
\IEEEauthorblockA{\footnotesize\texttt{\{xbalazia, sojka\}@fi.muni.cz}, \texttt{\{katerina.schindlerova, claudia.plant\}@univie.ac.at}}
\vspace{-16pt}
}
\date{}
\maketitle

\begin{abstract}
Which joint interactions in the human gait cycle can be used as biometric characteristics? Most current methods on gait recognition suffer from the lack of interpretability. We propose an interpretable feature representation of gait sequences by the graphical Granger causal inference. Gait sequence of a person in the standardized motion capture format, constituting a set of 3D joint spatial trajectories, is envisaged as a causal system of joints interacting in time. We apply the graphical Granger model (GGM) to obtain the so-called Granger causal graph among joints as a discriminative and visually interpretable representation of a person's gait. We evaluate eleven distance functions in the GGM feature space by established classification and class-separability evaluation metrics. Our experiments indicate that, depending on the metric, the most appropriate distance functions for the GGM are the total norm distance and the Ky-Fan 1-norm distance. Experiments also show that the GGM is able to detect the most discriminative joint interactions and that it outperforms five related interpretable models in correct classification rate and in Davies-Bouldin index. The proposed GGM model can serve as a complementary tool for gait analysis in kinesiology or for gait recognition in video surveillance.
\end{abstract}

\vspace{-6pt}
\section{Introduction}
\label{intro}

Human gait can be seen as a process in which joints of the corresponding skeleton interact in time and space. The movement of each joint represents a time series of spatio-temporal values. Such motion data were first collected and examined for gait recognition by Tanawongsuwan and Bobick~\cite{TB01} in 2001 to compose the gait features as four lower-body joint angle signals projected onto the walking plane. But the research has gone a long way since. Our colleagues have introduced various appearance models~\cite{YBS09}, relational features~\cite{MBS09}, geometric features~\cite{DMG14,APG15,KKMJ14}.

More recently, graphical representations~\cite{MMS13} and advanced machine learning methods~\cite{WGZW16,WBR16,LZWW15} have been brought in. Balazia~\etal~\cite{BS18} propose two sets of latent features learned by the maximum margin criterion and by a combination of principal component analysis and linear discriminant analysis, respectively. Kastaniotis~\etal~\cite{KTEF16,KTTEF15} fuse information from feature representations from both Euclidean and Riemannian spaces by mapping data in a reproducing kernel Hilbert space. All these methods, however, lack one thing in common: interpretability.

\begin{figure}[ht]
\centering
\vspace{-8pt}
\includegraphics[width=0.48\textwidth]{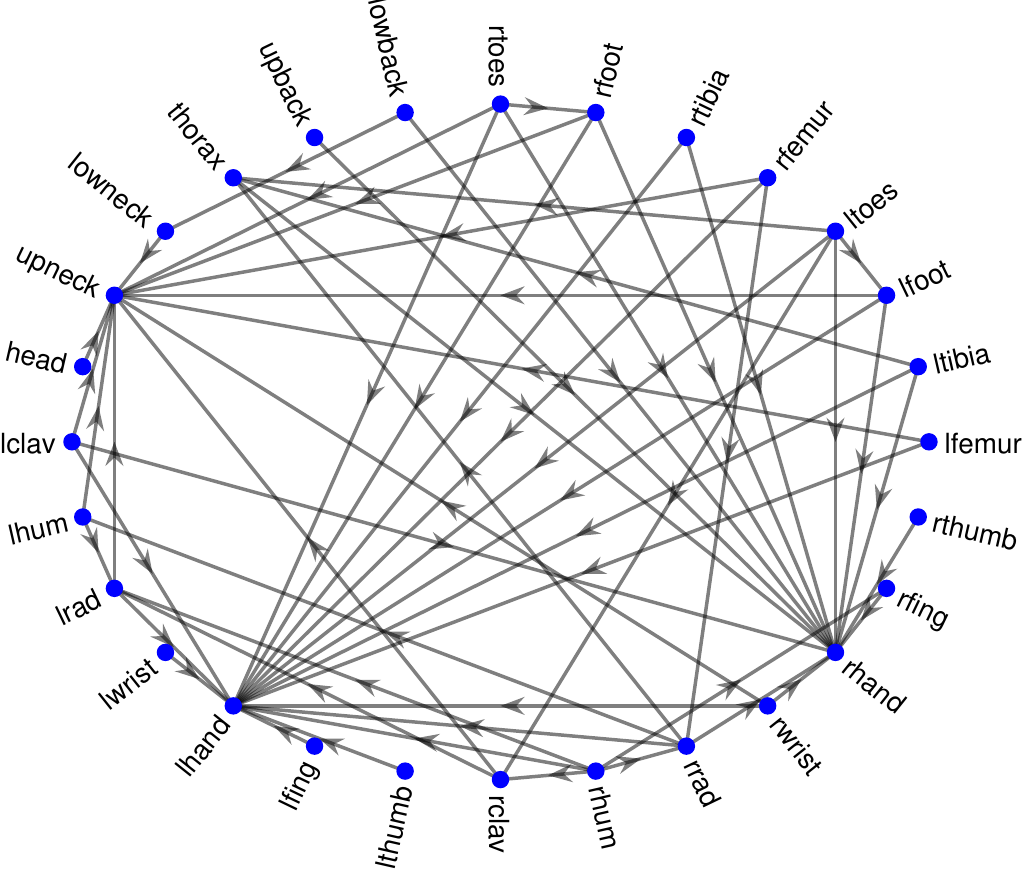}
\vspace{-8pt}
\caption{Visualization of the GGM gait feature of one gait sequence. Vertices denote joints and directed edges denote the Granger causal relations between them.}
\vspace{-10pt}
\label{f4}
\end{figure}

In this paper we ask: \emph{Which joint interactions in the human gait cycle can be used as biometric characteristics?} We approach this question by applying the Granger causality~\cite{granger1969} to compute a directed graph which expresses spatio-temporal interactions between body landmarks as temporal variables by the so-called graphical Granger models~(GGM)~\cite{arnold,hlavavckova2020heterogeneous}. Illustrated in Figure~\ref{f4}, the graph encodes the feature representation of one gait sequence, which we call the GGM gait feature. As of now, it is a conceptually unique model for visualizing and interpreting structured gait data and can serve as a tool for gait analysis in medical physiology or for gait recognition in video surveillance. Main contributions of this paper are:
\begin{itemize}[leftmargin=10pt,itemsep=0pt]
\item Construction of a graphical Granger model for identifying temporal interaction of joints by Granger causality. 
\item Investigation of eleven distance functions on how well they discriminate identities in the GGM feature space.
\item Experimental evaluation demonstrating that the GGM detects the most discriminative joint interactions and outperforms five related interpretable models.
\end{itemize}

\section{Motion Capture Data}
\label{mcd}
The proposed approach uses an advanced 3D motion acquisition technology that captures video clips of moving individuals and derives structural kinematic data. The format maintains an overall structure of the human body and holds estimated 3D positions of the main anatomical landmarks at a sequence of frames of synchronized and regular time intervals as the person moves. These so-called motion capture data~(MoCap) can be collected online by the \mbox{RGB-D} sensors such as Microsoft Kinect~\cite{kinect} or Vicon~\cite{vicon}.\looseness=-1

\begin{figure}[ht]
\centering
\vspace{-6pt}
\includegraphics[width=0.4\textwidth]{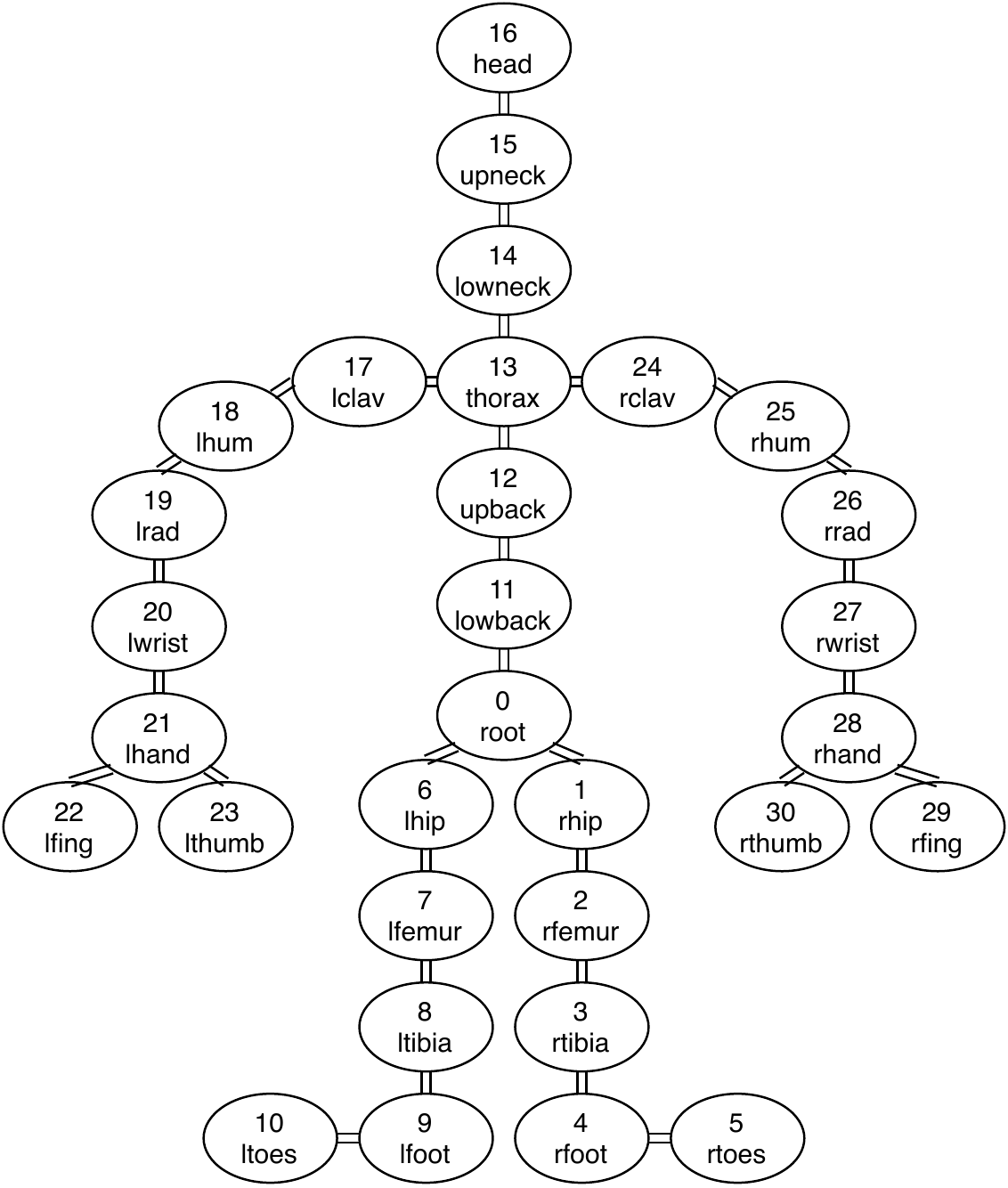}
\vspace{-7pt}
\caption{MoCap data. Skeleton is represented by a stick figure of 31 joints.}
\vspace{-6pt}
\label{f1}
\end{figure}

For a schematic visualization of MoCap, we typically use a simplified stick figure representing the human skeleton, a graph of joints connected by bones, as shown in Figure~\ref{f1}. These stick figures can be automatically recovered from body point spatial (Cartesian) coordinates in time as the person moves. The topology contains major body components (head, torso, hips, arms, thighs, knees and ankles) of proportional length, width and position.
The model is constructed with justifiable assumptions, only accounting for physiologically normal gait.



\section{Related Interpretable Models}
\label{sec_rel}

In order to reduce the high-dimensional MoCap data and to retain the discriminatory information at the same time, many research groups propose interpretable geometric features. These typically combine static body parameters~(bone lengths, person's height) with dynamic gait features such as step length, walk speed, joint angles and inter-joint distances, sometimes along with various statistics~(mean, standard deviation or local\slash global extremes) of their signals. Leveraging their temporal variations, these signals form gait templates and are compared by the dynamic time warping~(DTW) usually with one of the Levenshtein distances such as $L_1$~(Manhattan, CityBlock) or $L_2$~(Euclidean).

Jiang~\etal~\cite{JWZS15} extract 4 distances between joints as dynamic gait features. Ahmed~\etal~\cite{APG15} fuse 20 joint relative distances and 16 joint relative angles. Krzeszowski~\etal~\cite{KSKJW14} and Kwolek~\etal~\cite{KKMJ14} combine 26 pose attributes, such as bone angles (bone rotations around three axes), inter-joint distances, and the person's height. Their DTW-based baseline 1-NN classifier uses a distance function that measures differences in Euler angles, step length and height. Sedmidubsky~\etal~\cite{SVBZ12} concludes that only the two arm angle signals are discriminatory enough to be used for recognition.

One practice of these methods is to feed the extracted features into various statistical machine learning models, such as naïve Bayes or multilayer perceptron, which completely shroud any interpretable information. Another practice is to use the features while keeping their geometric properties, allowing for interpretation and so for evaluation. For the evaluations in this work, we follow the second practice and implement their features in geometric form.


\vspace{-4pt}
\section{Granger Causal Inference}
\label{sec_gr}

Since its introduction for bivariate case~\cite{granger1969}, Granger causality has been widely used for causal inference among temporal variables, e.g.~\cite{bressler2011wiener,seth2015granger}. The concept of Granger causality between two variables~$x$ and $y$ represented by two multivariate time series is defined as follows. Let $\bx^{1:n}=\left\{x^t|t=1,\ldots,n\right\}$ and $\by^{1:n}=\left\{y^t|t=1,\ldots,n\right\}$ be two time series up to time~$n$. Based on the following regression models, Granger causality between $\bx$ and $\by$ with lag $d=1$ is defined as
\vspace{-5pt}
\begin{align}
y^n & = \bB_1 \cdot \by^{1:(n-1)} + \bB_2 \cdot \bx^{1:(n-1)} + \varepsilon^n
\label{granger3}\\
y^n & = \bB_1 \cdot \by^{1:(n-1)} + \varepsilon^n
\label{granger4}
\end{align}
where $\varepsilon^t, t=1,\ldots,n$ are a Gaussian error time series with zero mean and $\bB_1$ and $\bB_2$ are matrices of coefficients for time series $\by^{1:n}$ and $\bx^{1:n}$, respectively. Here, $\bx$ is said to Granger-cause $\by$ if the first model in Eq.~\eqref{granger3} results in a significant improvement over the second model in Eq.~\eqref{granger4}. 

In the last decade, Granger causal inference among $p\!\ge\!3$ variables has been generalized in the form of GGMs by means of penalized regression models, e.g.~\cite{arnold,lozanospatial}, defined as follows. Given a multivariate series of observations $\left[x_i^t\right]_{t=1,\ldots,n, i=1,\ldots,p}$ where $n$ is the length of the time series and $p$ its dimension, let
\vspace{-4pt}
\begin{equation}\label{lag22}
\bX^\mathrm{lag}_{t,d} = \left[x_j^{t-k}\right]_{j=1,\ldots,p, k=1,\ldots,d, t=d+1,\dots,n}
\vspace{-4pt}
\end{equation}
for an integer $0<d<n$ denote a concatenated vector of all the lagged variables with maximal lag $d$ up to time $t$. The goal of the GGM is to compute a directed graph $G=(V,E)$ with vertices $V=\{1,\ldots,p\}$ as joints of the skeleton and directed edges~$E \subseteq V^2$ as existing causal directions between them. The edges $e_{ij} \in E$ correspond to the non-zero solutions of coefficients $\bbeta_i=\left[\beta_i^1\cdots\beta_i^p\right]$ of $p$ regression problems
\vspace{-4pt}
\begin{equation}\label{lagedreg}
\bx_i = \bX^\mathrm{lag}_{t,d}\bbeta_i, \quad i=1,\ldots,p.
\vspace{-4pt}
\end{equation}
Calculation of $\bbeta_i$ directly from Eq.~\eqref{lagedreg} through normal equations would lead to the form
\vspace{-4pt}
\begin{equation}
\bbeta_i = \left(\left(\bX^\mathrm{lag}_{t,d}\right)^{\top}\bX^\mathrm{lag}_{t,d}\right)^{-1}
\left(\bX^\mathrm{lag}_{t,d}\right)^{\top}\bx_i
\vspace{-4pt}
\end{equation}
although the above inverse might not exist as matrix $\left(\bX^\mathrm{lag}_{t,d}\right)^{\top} \bX^\mathrm{lag}_{t,d}$ is highly collinear due to its construction from the lagged variables.

Therefore, we use a lasso penalization method~\cite{lasso} allowing optimization with highly collinear matrices of covariates.
Lasso penalization with the GGM was first proposed by Arnold~\etal in~\cite{arnold} in the form
\vspace{-4pt}
\begin{equation}\label{penalized3}
\hat{\bbeta}_i = \argmin_{\bbeta_i}\sum_{t=d+1}^n\left|x_i^t-\bX^{lag}_{t,d}\bbeta_i\right|^2 + \lambda\left|\bbeta_i\right|.
\vspace{-4pt}
\end{equation}
The regularization parameter $\lambda>0$ controls the amount of shrinkage of the least squares and the variable selection, which, in moderation, can improve both prediction accuracy and interpretability. Cross-validation is usually used to select $\lambda$ within a given interval. The optimization of Eq.~\eqref{penalized3} can be calculated from the input $\bX$, lag $d$ and an upper bound for $\lambda$, $\lambda_\mathrm{max}$, for example by the coordinate descent~(CD) method or by the least-angle regression~(LARS) algorithm~\cite{lars}.

\textbf{Definition.} A Granger causal relation between two time series is defined based on the estimated coefficients $\hat{\bbeta_i}$. For a fixed lag $d>0$, the time series $\bx_j$ Granger-causes the time series $\bx_i$, denoted by $\bx_j \to \bx_i$, if at least one of the coefficients in the $j$-th column of $\hat{\bbeta_i}$ is non-zero.

Unfortunately, linear regression with lasso does not provide a unique general solution~\cite{zou2006adaptive}, i.e. it is not consistent so Granger causal conclusions from Eq.~\eqref{penalized3} can be spurious. To guarantee consistency, we use the GGM with the adaptive lasso. Moreover, in the following section we propose a generalization of the GGM for 3D time series.

\section{Graphical Granger Model for 3D Time Series}
\label{ggm3D}

In this section we construct a GGM for three-dimensional time series of MoCap data of human gait.
Assume a realistic model of the model of a human body with $p$~joints, such as in Figure~\ref{f1}, and a digitally captured motion sequence of length $n$~video frames. The tensor representation of a raw motion data sample has the form
\vspace{-4pt}
\begin{equation}\label{e1-0}
\bX = \begin{bmatrix}
x_1^1(\texttt{x}) & \cdots & x_1^n(\texttt{x})\\
x_1^1(\texttt{y}) & \cdots & x_1^n(\texttt{y})\\
x_1^1(\texttt{z}) & \cdots & x_1^n(\texttt{z})\\
\vdots & \ddots & \vdots\\
x_p^1(\texttt{x}) & \cdots & x_p^n(\texttt{x})\\
x_p^1(\texttt{y}) & \cdots & x_p^n(\texttt{y})\\
x_p^1(\texttt{z}) & \cdots & x_p^n(\texttt{z})\\
\end{bmatrix}
\end{equation}
of concatenated 3D spatial coordinates. Data in this format are considered the raw input for extracting the GGM gait feature. Since each of the $p$~time series in Eq.~\eqref{e1-0} is \mbox{3-dimensional} and one cannot apply the GGM from Eq.~\eqref{penalized3} directly, we order all dimensions of $\bx_i$ in one row\looseness=-1
\vspace{-5pt}
\begin{equation}\label{e1-1+}
\arraycolsep2pt
\bx_i=
\begin{bmatrix}
x_i^{d+1}(\texttt{x}) & x_i^{d+1}(\texttt{y}) & x_i^{d+1}(\texttt{z}) & \cdots & x_i^n(\texttt{x}) & x_i^n(\texttt{y}) & x_i^n(\texttt{z})
\end{bmatrix}.
\vspace{-2pt}
\end{equation}
Instead of using $\bX^{lag}_{t,d}$ depending on $t$ as in Eq.~\eqref{lagedreg}, we construct a fixed design matrix
\vspace{-4pt}
\begin{equation}\label{matrixX}
\arraycolsep0.7pt
\bX^\mathrm{lag}=
\begin{bmatrix}
x^{d}_1(\texttt{x}) & \cdots & x_1^1(\texttt{x}) & \cdots & x^{d}_p(\texttt{x}) & \cdots & x_p^1(\texttt{x}) \\
x^{d}_1(\texttt{y}) & \cdots & x_1^1(\texttt{y}) & \cdots & x^{d}_p(\texttt{y}) & \cdots & x_p^1(\texttt{y}) \\
x^{d}_1(\texttt{z}) & \cdots & x_1^1(\texttt{z}) & \cdots & x^{d}_p(\texttt{z}) & \cdots & x_p^1(\texttt{z}) \\
\vdots & \vdots & \vdots & \ddots & \vdots & \vdots & \vdots \\
x^{n-1}_1(\texttt{x}) & \cdots & x_1^{n-d+1}(\texttt{x}) & \cdots & x^{n-1}_p(\texttt{x}) & \cdots & x_p^{n-d+1}(\texttt{x}) \\
x^{n-1}_1(\texttt{y}) & \cdots & x_1^{n-d+1}(\texttt{y}) & \cdots & x^{n-1}_p(\texttt{y}) & \cdots & x_p^{n-d+1}(\texttt{y}) \\
x^{n-1}_1(\texttt{z}) & \cdots & x_1^{n-d+1}(\texttt{z}) & \cdots & x^{n-1}_p(\texttt{z}) & \cdots & x_p^{n-d+1}(\texttt{z}) \\
\end{bmatrix}
\vspace{-8pt}
\end{equation}
for the corresponding regression problem.

\textbf{Remark.} Vectors $\bX^\mathrm{lag}_{t,d}$ from Eq.~\eqref{lag22} are not rows of the matrix $\bX^\mathrm{lag}$. Instead, the matrix $\bX^\mathrm{lag}$ is constructed to allow the generalization of GGM problem, from one-dimensional $\bx_i$ resulting in a single row in $\bX^\mathrm{lag}$, to three dimensions. We see that $\bX_{t,d}^\mathrm{lag}\bbeta_i=\left(\bX^\mathrm{lag}\bbeta_i^{\top}\right)^t$ holds for each $t=d+1,\dots,n$ where $(\bv)^t$ denotes the $t$-th coordinate of a vector~$\bv$.

Now, substituting $\bX^\mathrm{lag}_{t,d}$ with $\bX^\mathrm{lag}$ and considering the 3D $\bx_i$ from Eq.~\eqref{e1-1+}, we obtain an equivalent problem to Eq.~\eqref{lagedreg},
\vspace{-5pt}
\begin{equation}\label{lagedreg2}
\bx_i^{\top} = \bX^\mathrm{lag}\bbeta_i^{\top}, \quad i=1,\dots,p.
\vspace{-4pt}
\end{equation}
Even though also Eq.~\eqref{lagedreg2} cannot be solved by normal equations due to the same invertibility issue as in Eq.~\eqref{lagedreg}, a penalized version of Eq.~\eqref{lagedreg2} can be solved. We use the adaptive lasso penalty which was introduced in~\cite{zou2006adaptive} for the general linear regression. The GGM problem has now the form
\vspace{-8pt}
\begin{equation}\label{penal}
\hat{\bbeta}_i = \argmin_{\bbeta_i} \sum_{t=d+1}^{n} \left\|\left(\bx_i^{\top}\right)^t-\left(\bX^\mathrm{lag}\bbeta_i^{\top}\right)^t\right\|_2^2 + \lambda \sum_{j=1}^p w_j\|\bbeta_j\|
\vspace{-2pt}
\end{equation}
for a given positive regularization parameter $\lambda \ge 0$ and with the weights
\vspace{-11pt}
\begin{equation}\label{penal2+}
w_j = \left\|\hat{\bbeta}_j^\mathrm{mle}\right\|^{-1}
\vspace{-4pt}
\end{equation}
as the total norm of the initial maximum likelihood~(ML) estimate $\hat{\bbeta}_j^\mathrm{mle}$ of parameters $\left[\bbeta_j\right]_{j=1,\ldots,p}$ which can be computed from the input $\bX$ using the iteratively reweighted least squares~(IRLS) algorithm~\cite{green}.

\textbf{Remark.} For a Gaussian regression with adaptive lasso, the solution to the Eq.~\eqref{penal} is unique~\cite{zou2006adaptive} and converges to the global optimum, that is, the problem is guaranteed to be consistent. 
In the experimental Section~\ref{exp-setup}, we provide setup specifications of this model on MoCap data.

\textbf{Remark.} A common way to select the lag parameter $d$ for the Granger model with lasso is to try these penalized regressions with various values of lag $d$ and track the Akaike information criterion~(AIC)~\cite{akaike1973} or the Bayes information criterion~(BIC)~\cite{schwarz1978} values.

Finally, the estimated values $\hat{\beta}_i^j$, $i,j=1,\ldots,p$ between the pairs of time series $\bx_i, i=1,\ldots,p$ serve as the edges in the resulting causal graph GGM if there exists a non-zero causality
\vspace{-6pt}
\begin{equation}
e_{ij}\in E\quad\Leftrightarrow\quad\exists t, 0<t<d:\left|\hat{\beta_i}^{j}(t)\right|>0.
\end{equation}

Figure~\ref{f4} illustrates the GGM gait feature from a gait sequence. Positioned on a circle in the counter-clockwise order, the $28$ vertices represent the joints described in Figure~\ref{f1}, with the exceptions of \texttt{root}, \texttt{lhip} and \texttt{rhip}, as in our data they have static spatial coordinates over time. The vertices are ordered systematically: left and right leg in the upper right segment, torso in the upper left segment, and left arm in the bottom left segment, and right arm in the bottom right segment. Directed edges between the vertices indicate the Granger causality of the source on the target. Specifically in this figure, one can observe causal directions from the legs to the upper torso. Furthermore, the joints with the highest in-going node degree are the left and right hand, which are the targets of mostly the leg joints. We interpret this observation with the fact that legs are the driving forces of a walk and that the arm movements are a consequence of the leg movements. In addition, the figure confirms the intuition that the upper body joint movements are not causal to the movements of remaining joints. To pose hypotheses from the point of view of physiology of musculoskeletal system or sport medicine is however beyond our expertise. We believe that the discriminative and interpretable GGM gait feature can provide a complementary tool to analyze gait of specific persons and extend the expert knowledge of researchers and medical professionals.

\vspace{-3pt}
\begin{algorithm}[ht]
\caption{Compute the GGM Gait Feature}
\textbf{Input:} $X=\left[x_i\right]$ from Eq.~\eqref{e1-0}, $d$, $\lambda_\mathrm{max}$\\
\textbf{Output:} GGM causality graph as adjacency matrix $A$
\begin{algorithmic}[1]
\Function{computeGGM}{$X,d,\lambda_\mathrm{max}$}
\State Compute design matrix $X^\mathrm{lag}$ using $X$ from Eq.~\eqref{matrixX}
\State Compute initial ML estimate $\hat{\beta}_j^\mathrm{mle}$ using $X$ by IRLS
\For {$x_i \in X$}
\State Compute $\hat{\beta}_i$ using $X$, $d$, $\lambda_\mathrm{max}$, $\hat{\beta}_j^\mathrm{mle}$ from Eq.~\eqref{penal} by LARS
\For {$\hat{\beta_i}^j$ columns of $\hat{\beta}_i$}
\State $A(j,i) \leftarrow 0$
\If {($\exists t, 0<t<d$ such that $\left|\hat{\beta_i}^{j}(t)\right|>0$)}
\State $A(j,i) \leftarrow 1$
\EndIf
\EndFor
\EndFor
\\\Return $A$
\EndFunction
\end{algorithmic}
\label{GGM_Algr}
\end{algorithm}

Algorithm~\ref{GGM_Algr} for computation of the GGM gait feature consists of these major steps: First, for $\bX$ we construct the lagged matrix $\bX^\mathrm{lag}$ of temporal variables as in Eq.~\eqref{matrixX}. Second, initial maximum likelihood estimates $\hat{\bbeta}_j^\mathrm{mle}$ of the parameters $\bbeta_j$ are computed for all $j=1,\ldots,p$ using the IRLS algorithm. Third, the estimates $\hat{\bbeta}_i, i=1,\ldots,p$, $\hat{\bbeta}_i\in\mathbb{R}^{1\times pd}$ are computed from matrix $\bX^\mathrm{lag}$, lag $d$, parameter $\lambda_\mathrm{max}$ and the initial ML estimate $\hat{\bbeta}_j^\mathrm{mle}$ by solving the problem in Eq.~\eqref{penal} with adaptive lasso. The optimization can be done by the coordinate descent or by the LARS algorithm. Fourth, the adjacency matrix $\bA$ of the causal graph GGM is constructed as follows: If at least one of the $d$~coefficients in the $j$-th row of $\hat{\bbeta}_i$ is non-zero, then we assign $A(j,i)=1$. The algorithm finally outputs $\bA$.
 

\vspace{-5pt}
\section{Distance Functions on Causal Graphs}
\label{distf}

A Granger causal graph is a directed graph that can be described by an adjacency matrix. Therefore, possible candidates for distance functions on such graphs include the well-known matrix norms. Assume a pair of $p\!\times\!p$ output causal adjacency matrices~$\bA=[a_{ij}]_{i,j=1}^p$ and~$\bA'=[a_{ij}']_{i,j=1}^p$. We assess the following distance functions $\delta\left(\bA,\bA'\right)$:
\begin{description}[leftmargin=0pt,noitemsep]
\item{\bf Distance functions defined over vector norms}
\begin{itemize}[leftmargin=0pt,noitemsep]
\item{\it Total (absolute) norm distance:} $\delta_T\left(\bA,\bA'\right)=\|\bA-\bA'\|$ with $\|\bB\|=\sum_{i,j=1}^p|b_{ij}|$ as the total (absolute) norm of $\bB$.
\item{\it Frobenius norm distance:} $\delta_F\left(\bA,\bA'\right)=\|\bA-\bA'\|_2$ with $\|\bB\|_2=\sqrt{\sum_{i,j=1}^p\left(b_{ij}\right)\!^2}$ as the Frobenius norm of $\bB$.
\item {\it Max norm distance:} $\delta_M\left(\bA,\bA'\right)=p\cdot\|\bA-\bA'\|_M$ with $\|\bB\|_M=\max_{i,j=1,\dots,p}|b_{ij}|$ as the max norm of $\bB$.
\item{\it Jaccard distance:} $\delta_J\left(\bA,\bA'\right)=\frac{\|\min(\bA,\bA')\|_2}{\|\max(\bA,\bA')\|_2}$ with $\min(\cdot,\cdot)$ and $\max(\cdot,\cdot)$ are matrices of element-wise minima and maxima, respectively.
\item{\it Hamming distance:} $\delta_H\!\!\left(\bA,\bA'\right)\!=\!\frac{\|\max(\bA,\bA')\|_2\!-\!\|\min(\bA,\bA')\|_2}{n(n-1)}$ with $\min(\cdot,\cdot)$ and $\max(\cdot,\cdot)$ as above.
\end{itemize}
\item{\bf Distance functions defined by operator norms}
\begin{itemize}[leftmargin=0pt,noitemsep]
\item{\it Absolute row sum norm distance:}
$\delta_\infty\!\left(\bA,\bA'\right)=\|\bA-\bA'\|_\infty$
with $\|\bB\|_\infty=\max_{i=1,\dots,p}\sum_{j=1}^p|b_{ij}|$ as the absolute row sum norm of $\bB$.
\item{\it Absolute column sum norm distance:} $\delta_1\!\!\left(\bA,\bA'\right)\!\!=\!\!\|\bA\!\!-\!\!\bA'\|_1$ with $\|\bB\|_1=\max_{j=1,\dots,p} \sum_{i=1}^p|b_{ij}|$ is the absolute column sum norm of $\bB$.
\item{\it Spectral norm distance:} $\delta_S\left(\bA,\bA'\right)=\|\bA-\bA'\|_S$
with $\|\bB\|_{S}=\sqrt{\lambda_1\left(\bB^{\top} \bB\right)}$ as the spectral norm of $\bB$ and $\lambda_k(\bB')$ as the $k$-th largest eigenvalue of $\bB'$.
\end{itemize}
\item{\bf Distance functions defined by singular values}
\begin{itemize}[leftmargin=0pt,noitemsep]
\item{\it Ky-Fan $k$-norm distance:} $\delta_{\mathit{KF}(k)}\left(\bA,\bA'\right)=\sum_{i=1}^k\sigma_i$ with $\sigma_i$ as the $i$-th singular value of $\left|\bA-\bA'\right|$.
\item{\it Hilbert-Schmidt norm distance:} $\delta_{\mathit{HS}}\left(\bA,\bA'\right)=\sqrt{\sum_{i=1}^r\sigma_i^2}$ with $\sigma_i$ as the $i$-th singular value of $\left|\bA-\bA'\right|$ and $r=\mathit{rank}\left(|\bA-\bA'|\right)$.
\end{itemize}
\item{\bf Distance functions induced by weighted vector norms}
\begin{itemize}[leftmargin=0pt,noitemsep]
\item {\it Mahalanobis distance:} a weighted Euclidean distance $\delta_M\left(\bA,\bA'\right)=\sqrt{\left\|(\bA-\bA')^{\top}\bSigma_T^{-1}(\bA-\bA')\right\|}$ with $\bSigma_T$ as the total scatter matrix of all adjacency matrices and $\|\bB\|=\sum_{i,j=1}^p|b_{ij}|$ as the total norm of $\bB$.
\end{itemize}
\end{description}



\section{Evaluation}
\label{exp}

\subsection{Data}

We have extracted $302$~gait samples of $16$~identities from the CMU MoCap dataset~\cite{CMU03} recorded with the optical marker-based system Vicon~\cite{vicon}. Data are stored in the standard ASF/AMC data format where the ASF files contain each person's static 31-joint skeleton parameters and the AMC files describe bone rotational data during motion. 3D joint coordinates are calculated using bone lengths and rotations. The second dataset is KinectUNITO~\cite{GBGL13,GGLB14} acquired with the optical marker-free system Microsoft Kinect~\cite{kinect}. Kinect provides a 20-joint skeleton and this dataset contains $400$~gait samples of $20$~identities.

To ensure skeleton invariance, we use one prototypical skeleton as the mean of all skeletons in the dataset. To ensure translation and rotation invariance, the center of the coordinate system $[0,0,0]$ is translated to the \texttt{root} joint and the axes are rotated according to the walker's perspective: X~axis is from right (negative) to left (positive), Y~axis is from down (negative) to up (positive), and Z~axis is from back (negative) to front (positive). For each skeleton, we consider $p=28$ of the $31$~joints modeled in Figure~\ref{f1} of CMU MoCap and $p=17$ of the $20$~joints of KinectUNITO, with the three exceptions of \texttt{root}, \texttt{lhip} and \texttt{rhip}, as they are static over time and without an impact on causality. As a repetitive unit pattern of walking, each complete gait cycle is used as a biometric sample.
We further set a fixed length $n=156$ frames for each gait cycle, which is selected as the average gait cycle length. 

\vspace{-3pt}
\subsection{Optimal Configuration of GGM}
\label{exp-setup}

Before applying the GGM approach to the given dataset, we proved its feasibility in Section~\ref{ggm3D}. By statistical testing we confirmed Gaussianity and stationarity of all time series, so the GGM method is feasible for causal inference among the joints.

An appropriate lag $d$ for time series in Eq.~\eqref{penal} can be calculated by AIC or BIC assuming that the degrees of freedom are equal to the number of non-zero parameters, which is only known to be true for the lasso penalty~\cite{zou2006adaptive} but it is unknown for adaptive lasso. To select the lag, we followed the observation of Behzadi~\etal~\cite{behzadi} that varying the lag parameter from $1$ to $50$ has a negligible impact on the performance of GGM with adaptive lasso. In our experiments we therefore consider the lag $d=1$.

For computation of Eq.~\eqref{penal} via coordinate descent, we used adaptive lasso from package~\cite{penalized} and selected $\lambda$ from interval $(0,\lambda_\mathrm{max}]$ with $\lambda_\mathrm{max}=5$. We found $\hat{\bbeta}_i$ and $\lambda$ using 5-fold cross-validation with respect to $\lambda$.

The final GGM parameter to be optimized is the distance function. We evaluate all distance functions in Section~\ref{distf} by correct classification rate~(CCR) and clustering criteria Davies-Bouldin index~(DBI) and Dunn index~(DI).
\subsection{Results}

We provide the results on an evaluation of GGM's distance function as an ablation study, a comparison against related models, and an analysis of discriminativeness of all pairs of joint interactions. Additional analyses of lag value, dataset size and interpretability of related graphical models are provided in the supplementary material.

\vspace{-1pt}
\subsubsection{Distance Function}
\label{sec_dist}

The goal was to find a distance function on adjacency matrices of causal graphs optimizing all CCR, DBI and DI at the same time. Our experimental results on both datasets CMU MoCap and KinectUNITO are reported in Table~\ref{tx}.
The main observation is that the total norm and the Ky-Fan 1-norm outperform other distance functions in CCR and DBI by a noticeable margin.

\begin{table}[ht]
\centering
\tabcolsep2.5pt
\vspace{-10pt}
\caption{GGM with 11 distance functions evaluated on CCR, DBI, DI.}
\vspace{-8pt}
\begin{tabular}{|r|ccc|ccc|}
\hline
dataset & \multicolumn{3}{c|}{CMU MoCap~\cite{CMU03}} & \multicolumn{3}{c|}{KinectUNITO~\cite{GBGL13}} \\\hline
distance function & CCR $\uparrow$ & DBI $\downarrow$ & DI $\uparrow$ & CCR $\uparrow$ & DBI $\downarrow$ & DI $\uparrow$ \\\hline
total norm & \textbf{0.9250} & 0.7051 & 1.2057 & \textbf{0.8541} & 0.7523 & 1.7266 \\
Frobenius norm & 0.8059 & 0.5347 & 1.6013 & 0.7626 & 0.5644 & 1.6903 \\
max norm & 0.7556 & 0.5150 & 1.7429 & 0.8032 & 0.6044 & 1.5359 \\
Jaccard & 0.6938 & 0.9401 & 0.9043 & 0.7523 & 1.1240 & 1.3402 \\
Hamming & 0.6850 & 0.7129 & 1.3611 & 0.7626 & 0.8634 & 1.4521 \\
abs. row sum norm & 0.7873 & 0.6441 & 1.3608 & 0.7217 & 0.6251 & 1.7114 \\
abs. col. sum norm & 0.7752 & 0.5062 & 1.7954 & 0.7338 & 0.6287 & 1.7497 \\
spectral norm & 0.7727 & 0.5551 & 1.6330 & 0.6710 & 0.6598 & 1.6607 \\
Ky-Fan 1-norm & 0.7861 & \textbf{0.4981} & \textbf{1.8393} & 0.6934 & \textbf{0.5267} & 1.7844 \\
Hilbert-Schmidt norm & 0.8390 & 0.5793 & 1.5488 & 0.8206 & 0.5901 & 1.4889 \\
Mahalanobis & 0.8059 & 0.5672 & 1.5909 & 0.8133 & 0.5772 & \textbf{1.7905} \\\hline
\end{tabular}
\vspace{-10pt}
\label{tx}
\end{table}



\vspace{-1pt}
\subsubsection{Comparison to Related Interpretable Models}

Our method is compared to all interpretable models from Section~\ref{sec_rel}.
All methods have been implemented with the highest scoring configuration reported in the respective papers. Results in terms of CCR, DBI and DI are shown in Table~\ref{tr}. Our GGM with total norm distance scored the highest CCR, the one with Ky-Fan 1-norm distance has the lowest DBI and the one with Mahalanobis distance falls second/third by DI behind the model of Kwolek~\etal~\cite{KKMJ14}. Overall, we interpret the comparative evaluation as that the GGM gait feature is an interpretable MoCap-based gait recognition model of a high discrimination ability.

\begin{table}[ht]
\centering
\tabcolsep3pt
\vspace{-10pt}
\caption{GGM and 5 interpretable models evaluated on CCR, DBI, DI.}
\vspace{-8pt}
\begin{tabular}{|r|ccc|ccc|}
\hline
dataset & \multicolumn{3}{c|}{CMU MoCap~\cite{CMU03}} & \multicolumn{3}{c|}{KinectUNITO~\cite{GBGL13}} \\\hline
model & CCR $\uparrow$ & DBI $\downarrow$ & DI $\uparrow$ & CCR $\uparrow$ & DBI $\downarrow$ & DI $\uparrow$ \\\hline
Ahmed~\cite{APG15} & 0.7705 & 0.6617 & 1.7051 & 0.7273 & 0.7219 & 1.7225 \\
Jiang~\cite{JWZS15} & 0.7149 & 0.7387 & 1.5104 & 0.6948 & 0.8345 & 1.6372 \\
Krzeszowski~\cite{KSKJW14} & 0.8258 & 0.6260 & 1.8033 & 0.8037 & 0.6031 & 1.8158 \\
Kwolek~\cite{KKMJ14} & 0.9046 & 0.5639 & \textbf{1.8652} & 0.8520 & 0.7692 & \textbf{1.8627} \\
Sedmidubsky~\cite{SVBZ12} & 0.6854 & 0.9410 & 1.2757 & 0.7609 & 0.6914 & 1.6630 \\
\textbf{GGM (proposed)} & \textbf{0.9250} & \textbf{0.4981} & 1.8393 & \textbf{0.8541} & \textbf{0.5267} & 1.7905 \\\hline
\end{tabular}
\vspace{-10pt}
\label{tr}
\end{table}

\subsubsection{Joint Interactions}

\begin{figure}[t!]
\centering
\vspace{-5pt}
\includegraphics[width=0.49\textwidth]{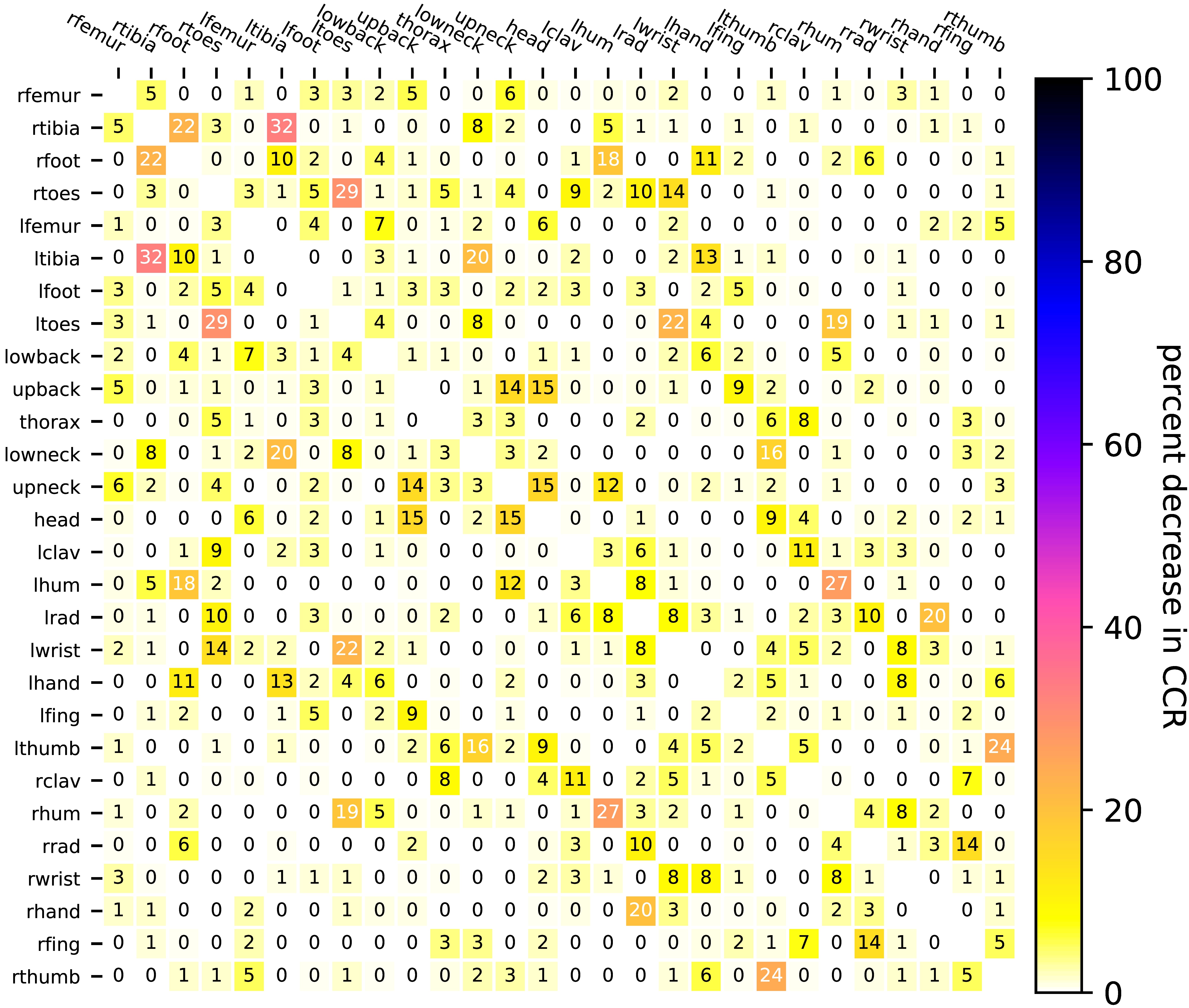}
\includegraphics[width=0.49\textwidth]{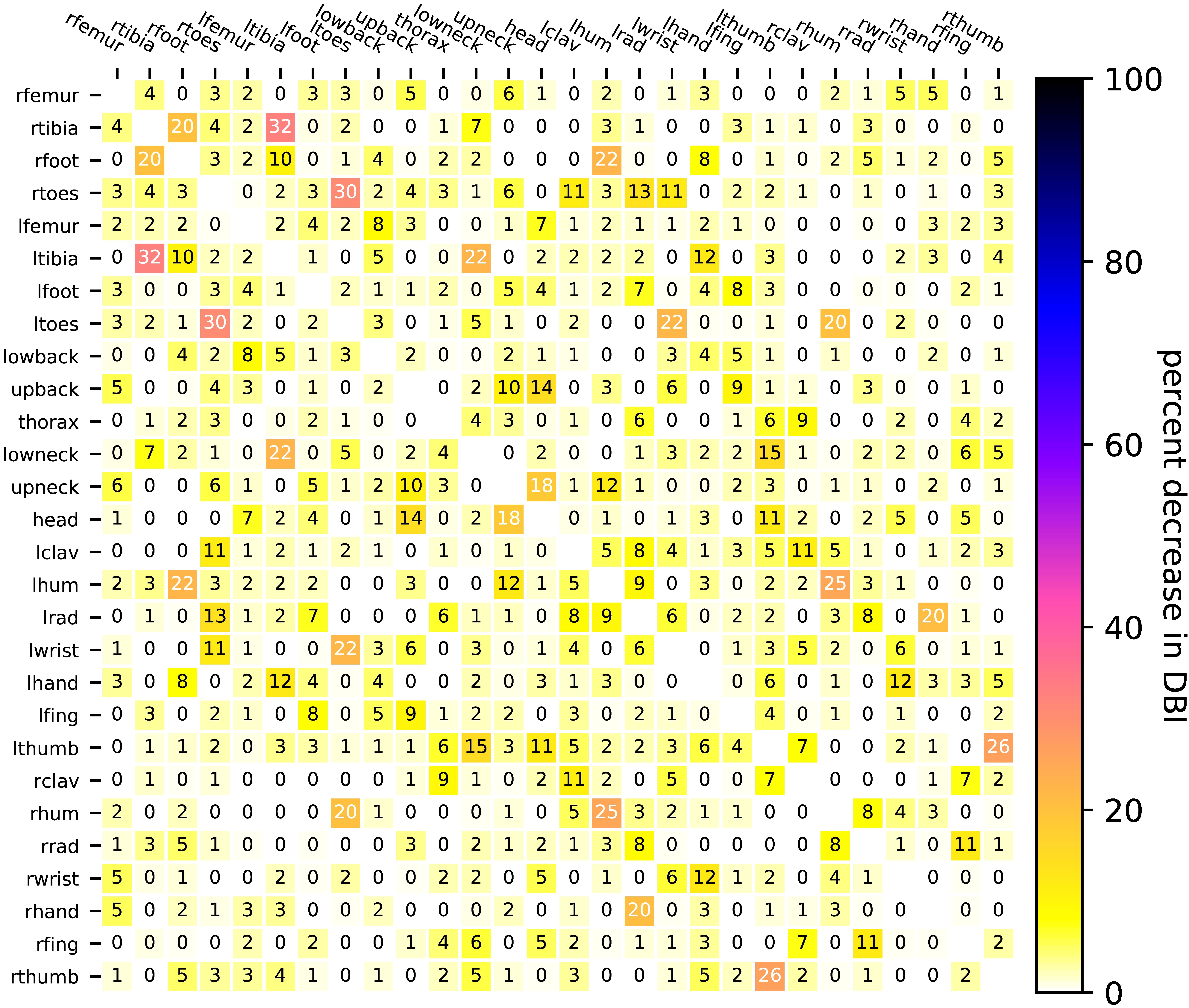}
\includegraphics[width=0.49\textwidth]{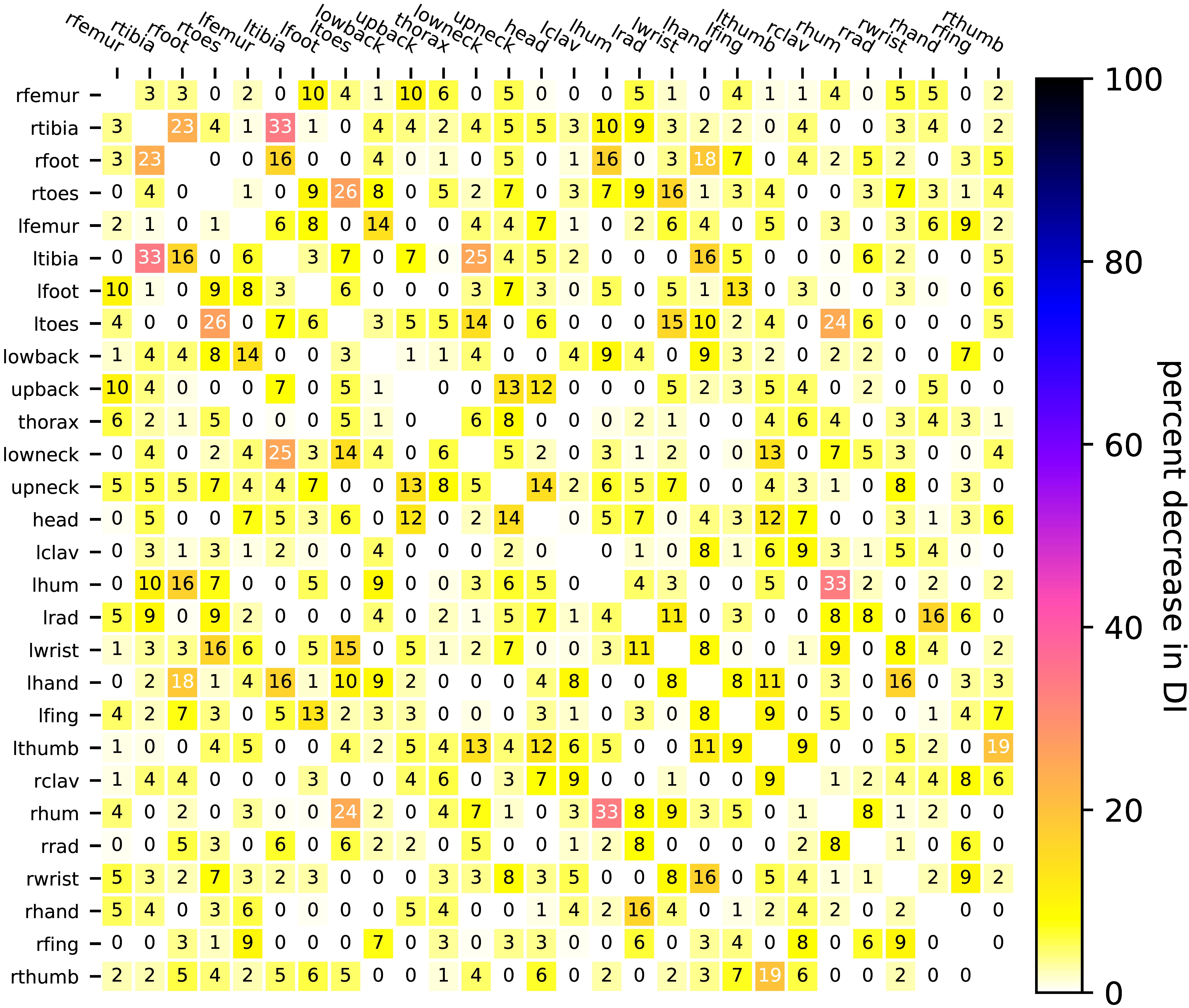}
\vspace{-20pt}
\caption{Percentual decrease in discrimination ability by CCR~(top), DBI~(middle) and DI~(bottom) of a large number of GGMs after removing the causality between a joint in row and a joint in column.}
\vspace{-25pt}
\label{f6}
\end{figure}

In this part we answer the question from the introduction about the discriminative ability of all joint interactions. The proposed experiment compares a large series of GGMs, each modified by removing one particular joint pair. Discriminative ability of an individual joint interaction is quantified as the percentual decrease in the evaluation metrics: lower CCR, higher DBI and lower DI. As the distance function, we select the optimal one in each metric given by the results in Table~\ref{tx}. Figure~\ref{f6} shows the percentual decrease in CCR, DBI and DI of GGMs modified by removing all pairs joint causalities, one at a time. For illustration, the value in the CCR matrix between \texttt{ltibia--rtibia} is 32, which means that disregarding the causal edge between ankles lowers the CCR by 32\% compared to the complete model. Joint pairs of the highest discriminative causalities are ankles (\texttt{ltibia--rtibia}), toes (\texttt{ltoes--rtoes}) and elbows (\texttt{lhum--rhum}). Another observation is a high variability in discriminativess across joint pairs. And the final remark is that CCR is the least sensitive, while DI is the most.

\vspace{-5pt}
\section{Conclusion}
\label{concl}


We proposed a new feature representation of gait sequences in the standardized MoCap format by Granger causal inference, called the GGM gait feature, to encode temporal interactions of joints during walk. To the best of our knowledge, this is the first work applying the graphical Granger causality to feature representation of gait.

To define the similarity of GGM gait features of different persons, we investigated eleven distance functions on their separability of the identity classes by three evaluation metrics CCR, DBI, and DI. Our experiments on CMU MoCap and KinectUNITO indicate that the most effective distance functions for GGM gait features are the total norm in terms of CCR and the Ky-Fan 1-norm distance by DBI and DI. We further discovered that the most discriminatory interactions are those between ankles, toes, and elbows. Finally, we compared the optimal configuration GGM against five related interpretable models and obtained the highest CCR, the lowest DBI and the second/third highest~DI.


In our future research, we proceed with analyzing the impact of additional skeleton schemes of OpenPose~\cite{openpose}, AlphaPose~\cite{alphapose} and LCRNet~\cite{lcrnet}, which consider fewer joints. We believe that pathways for improvement can be established by advancing deep neural networks designed on principles of interpretable or explainable AI.


\vspace{-5pt}
\section*{Acknowledgment}
\footnotesize
This work was supported by the French National Research Agency project ANR-15-IDEX-01 and by the Czech Science Foundation project GA19-16066S. Data used in this project were created with funding from NSF EIA-0196217 and were obtained from \url{http://mocap.cs.cmu.edu}. Our MATLAB code for GGM gait feature extraction is available at \url{https://dm.cs.univie.ac.at/research/downloads/} and our extracted MoCap database and evaluation framework in Java at \url{https://gait.fi.muni.cz}.


\normalsize
\bibliographystyle{abbrv}
\bibliography{ref}

\begin{thebibliography}{10}

\bibitem{APG15}
F.~Ahmed, P.~P. Paul, and M.~L. Gavrilova.
\newblock {DTW-Based Kernel and Rank-Level Fusion for 3D Gait Recognition using
  Kinect}.
\newblock {\em {The Visual Computer}}, 31(6):915--924, 2015.

\bibitem{akaike1973}
H.~Akaike.
\newblock {Information Theory and an Extension of the Maximum Likelihood
  Principle}.
\newblock In {\em 2nd International Symposium on Information Theory,
  Tsahkadsor, Armenia, USSR, September 2--8, 1971}, pages 267--281, Budapest,
  Hungary, 1973. Akademiai Kiado.

\bibitem{arnold}
A.~Arnold, Y.~Liu, and N.~Abe.
\newblock {Temporal Causal Modeling with Graphical Granger Methods}.
\newblock In {\em ACM SIGKDD}, KDD\,'07, pages 66--75, 2007.

\bibitem{BS18}
M.~Balazia and P.~Sojka.
\newblock {Gait Recognition from Motion Capture Data}.
\newblock {\em ACM Trans. Mult. Comp. Commun. Appl.}, 14(1s):22:1--22:18, 2018.

\bibitem{behzadi}
S.~Behzadi, K.~Hlav{\'a}{\v{c}}kov{\'a}-Schindler, and C.~Plant.
\newblock Granger causality for heterogeneous processes.
\newblock In {\em PAKDD}, pages 463--475. Springer, 2019.

\bibitem{bressler2011wiener}
S.~Bressler and A.~Seth.
\newblock Wiener-{G}ranger causality: {A} well established methodology.
\newblock {\em Neuroimage}, 58(2):323--329, 2011.

\bibitem{openpose}
Z.~Cao, G.~Hidalgo, T.~Simon, S.~Wei, and Y.~Sheikh.
\newblock Openpose: Realtime multi-person 2d pose estimation using part
  affinity fields.
\newblock {\em IEEE Transactions on Pattern Analysis \& Machine Intelligence},
  43(01):172--186, jan 2021.

\bibitem{CMU03}
{CMU Graphics Lab}.
\newblock {MoCap Database}, 2003.
\newblock Available at \url{http://mocap.cs.cmu.edu}, last access 11 Oct 2021.

\bibitem{DMG14}
B.~Dikovski, G.~Madjarov, and D.~Gjorgjevikj.
\newblock {Evaluation of Different Feature Sets for Gait Recognition using
  Skeletal Data from Kinect}.
\newblock In {\em {ICTEM}}, pages 1304--1308. IEEE, 2014.

\bibitem{lars}
B.~Efron, T.~Hastie, I.~Johnstone, and R.~Tibshirani.
\newblock {Least Angle Regression}.
\newblock {\em The Annals of Statistics}, 32(2):407--499, 2004.

\bibitem{alphapose}
H.-S. Fang, S.~Xie, Y.-W. Tai, and C.~Lu.
\newblock Rmpe: Regional multi-person pose estimation.
\newblock In {\em ICCV}, 2017.

\bibitem{GBGL13}
E.~Gianaria, N.~Balossino, M.~Grangetto, and M.~Lucenteforte.
\newblock Gait characterization using dynamic skeleton acquisition.
\newblock In {\em Multimedia Signal Processing (MMSP), 2013 IEEE 15th
  International Workshop on}, pages 440--445, Sept 2013.

\bibitem{GGLB14}
E.~Gianaria, M.~Grangetto, M.~Lucenteforte, and N.~Balossino.
\newblock Human classification using gait features.
\newblock In V.~Cantoni, D.~Dimov, and M.~Tistarelli, editors, {\em Biometric
  Authentication}, Lecture Notes in Computer Science, pages 16--27. Springer
  International Publishing, 2014.

\bibitem{granger1969}
C.~Granger.
\newblock {Investigating Causal Relations by Econometric Models and
  Cross-spectral Methods}.
\newblock {\em Econometrica}, 37(3):424--438, 1969.

\bibitem{green}
P.~Green.
\newblock {Iteratively Reweighted Least Squares for Maximum Likelihood
  Estimation, and Some Robust and Resistant Alternatives}.
\newblock {\em Journal of the Royal Statistical Society: Series B
  (Methodological)}, 46(2):149--170, 1984.

\bibitem{hlavavckova2020heterogeneous}
K.~Hlav{\'a}{\v{c}}kov{\'a}-Schindler and C.~Plant.
\newblock Heterogeneous graphical {G}ranger causality by minimum message
  length.
\newblock {\em Entropy}, 22(12):1400, 2020.

\bibitem{JWZS15}
S.~Jiang, Y.~Wang, Y.~Zhang, and J.~Sun.
\newblock {Real Time Gait Recognition System Based on Kinect Skeleton Feature}.
\newblock In {\em {ACCV-W}}, volume 9008, pages 46--57. Springer, 2015.

\bibitem{KTEF16}
D.~Kastaniotis, I.~Theodorakopoulos, G.~Economou, and S.~Fotopoulos.
\newblock {Gait Based Recognition via Fusing Information from Euclidean and
  Riemannian Manifolds}.
\newblock {\em {Pattern Recognition Letters}}, 84:245--251, 2016.

\bibitem{KTTEF15}
D.~Kastaniotis, I.~Theodorakopoulos, C.~Theoharatos, G.~Economou, and
  S.~Fotopoulos.
\newblock {A Framework for Gait-Based Recognition using Kinect}.
\newblock {\em {Pattern Recognition Letters}}, 68(2):327--335, 2015.
\newblock Special Issue on Soft Biometrics.

\bibitem{kinect}
Kinect.
\newblock
  \url{https://docs.microsoft.com/en-us/azure/kinect-dk/hardware-specification}.

\bibitem{KSKJW14}
T.~Krzeszowski, A.~\'{S}wito\'{n}ski, B.~Kwolek, H.~Josi\'{n}ski, and
  K.~Wojciech\'{o}wski.
\newblock {DTW-Based Gait Recognition from Recovered 3-D Joint Angles and
  Inter-Ankle Distance}.
\newblock In {\em {ICCVG}}, volume 8671 of {\em LNCS}, pages 356--363.
  Springer, 2014.

\bibitem{KKMJ14}
B.~Kwolek, T.~Krzeszowski, A.~Michalczuk, and H.~Josi\'{n}ski.
\newblock {3D Gait Recognition using Spatio-Temporal Motion Descriptors}.
\newblock In {\em {ACIIDS}}, pages 595--604. Springer, 2014.

\bibitem{LZWW15}
Z.~Liu, Z.~Zhang, Q.~Wu, and Y.~Wang.
\newblock {Enhancing Person Re-Identification by Integrating Gait Biometric}.
\newblock In {\em {Asian CCV}}, pages 35--45. Springer, 2015.

\bibitem{lozanospatial}
A.~Lozano, H.~Li, A.~Niculescu-Mizil, Y.~Liu, C.~Perlich, J.~Hosking, and
  N.~Abe.
\newblock Spatial-temporal causal modeling for climate change attribution.
\newblock In {\em ACM SIGKDD}, pages 587--596, 2009.

\bibitem{penalized}
W.~H. McIlhagga.
\newblock {penalized: A MATLAB Toolbox for Fitting Generalized Linear Models
  with Penalties}.
\newblock {\em Journal of Stat. Software}, 72(6):1--21, 2016.

\bibitem{MMS13}
M.~Milovanovic, M.~Minovic, and D.~Starcevic.
\newblock {Walking in Colors: Human Gait Recognition using Kinect and CBIR}.
\newblock {\em {Proc. of IEEE MultiMedia}}, 20(4):28--36, Oct 2013.

\bibitem{MBS09}
M.~M\"{u}ller, A.~Baak, and H.-P. Seidel.
\newblock {Efficient and Robust Annotation of Motion Capture Data}.
\newblock In {\em {ACM SIGGRAPH/Eurographics}}, SCA '09, pages 17--26, 2009.

\bibitem{lcrnet}
G.~Rogez, P.~Weinzaepfel, and C.~Schmid.
\newblock {LCR-Net++: Multi-person 2D and 3D Pose Detection in Natural Images}.
\newblock {\em {IEEE Transactions on Pattern Analysis and Machine
  Intelligence}}, 2019.

\bibitem{schwarz1978}
G.~Schwarz.
\newblock {Estimating the Dimension of a Model}.
\newblock {\em The Annals of Statistics}, 6(2):461--464, Mar. 1978.

\bibitem{SVBZ12}
J.~Sedmidubsky, J.~Valcik, M.~Balazia, and P.~Zezula.
\newblock {Gait Recognition Based on Normalized Walk Cycles}.
\newblock In {\em {Advances in Visual Computing}}, LNCS, pages 11--20.
  Springer, 2012.

\bibitem{seth2015granger}
A.~Seth, A.~Barrett, and L.~Barnett.
\newblock Granger causality analysis in neuroscience and neuroimaging.
\newblock {\em Journal of Neuroscience}, 35(8):3293--3297, 2015.

\bibitem{TB01}
R.~Tanawongsuwan and A.~Bobick.
\newblock {Gait Recognition from Time-Normalized Joint-Angle Trajectories in
  the Walking Plane}.
\newblock In {\em {CVPR}}, volume~2, pages 726--731. IEEE, 2001.

\bibitem{lasso}
R.~Tibshirani.
\newblock {Regression Shrinkage and Selection via the Lasso}.
\newblock {\em Journal of the Royal Statistical Society: Series B
  (Methodological)}, 58(1):267--288, 1996.

\bibitem{vicon}
Vicon.
\newblock \url{http://www.vicon-security.com/}.

\bibitem{WGZW16}
T.~Wang, S.~Gong, X.~Zhu, and S.~Wang.
\newblock {Person Re-Identification by Discriminative Selection in Video
  Ranking}.
\newblock {\em {IEEE Trans. on PAMI}}, 38(12):2501--2514, 2016.

\bibitem{WBR16}
T.~Wolf, M.~Babaee, and G.~Rigoll.
\newblock {Multi-View Gait Recognition using 3D Convolutional Neural Networks}.
\newblock In {\em {2016 IEEE International Conference on Image Processing
  (ICIP)}}, pages 4165--4169, Sept 2016.

\bibitem{YBS09}
K.~Yamauchi, B.~Bhanu, and H.~Saito.
\newblock {Recognition of Walking Humans in 3D}.
\newblock In {\em { CVPRW}}, pages 45--52, 2009.

\bibitem{zou2006adaptive}
H.~Zou.
\newblock {The Adaptive Lasso and Its Oracle Properties}.
\newblock {\em Journal of the Am. Stat. Ass.}, 101(476):1418--1429, 2006.

\end{thebibliography}
\end{document}